%% file: arxiv_preprint.tex
\documentclass{article}
\usepackage{arxiv}

\usepackage{amssymb}
\usepackage{amsmath}
 
\usepackage{hyperref}
\usepackage{mathtools}
\usepackage[capitalize,noabbrev]{cleveref}

\usepackage{microtype}
\usepackage{graphicx}
\usepackage[numbers]{natbib}
\usepackage{subcaption}
\usepackage{booktabs}
\usepackage[textsize=tiny]{todonotes}
\usepackage{float}
\usepackage{natbib} 

\input{macrosetup}

\title{Gauss-Newton Natural Gradient Descent for \\ Shape Learning}

\newcommand{\auth}[2]{#1$^{#2}$}
\newcommand{\affil}[2]{$^{#1}$ #2\\}

\author{
  \auth{James King}{1}  \qquad
  \auth{Arturs Berzins}{2}  \qquad
  \auth{Siddhartha Mishra}{1,3}  \qquad
  \auth{Marius Zeinhofer}{1} \\
  \\
  \affil{1}{Seminar for Applied Mathematics (SAM), D-Math, ETH Zürich, Switzerland}
  \affil{2}{Institute for Machine Learning, JKU Linz, Austria}
  \affil{3}{ETH AI Center, ETH Zürich, Switzerland}
}

\begin{document}
\maketitle

\begin{abstract} 
We explore the use of the Gauss-Newton method for optimization in shape learning, including implicit neural surfaces and geometry-informed neural networks. The method addresses key challenges in shape learning, such as the ill-conditioning of the underlying differential constraints and the mismatch between the optimization problem in parameter space and the function space where the problem is naturally posed.
This leads to significantly faster and more stable convergence than standard first-order methods, while also requiring far fewer iterations.
Experiments across benchmark shape optimization tasks demonstrate that the Gauss-Newton method consistently improves both training speed and final solution accuracy. 
\end{abstract}






\section{Introduction}

Shape learning methods, such as neural fields (NFs) \cite{Sitzmann2020ImplicitNR, essakine2025standimplicitneuralrepresentations} or implicit neural shapes (INSs) \cite{park2019deepsdflearningcontinuoussigned, chen2019learningimplicitfieldsgenerative}, are designed to learn geometric structures from a mixture of sparse data and geometric constraints. These approaches use neural networks to represent complex shapes and surfaces as (sub-)level sets of learned functions.
Shape learning methods offer a powerful way to represent and generate intricate geometries. Moreover, they are highly memory-efficient and inherently differentiable, making them suitable for applications in computer graphics, shape optimization, and generative design, where precise geometric control is essential. \\
A key aspect of shape learning is the incorporation of differential constraints into the learning process. These constraints, involving curvature, surface normals, and other geometric properties, are essential for ensuring that the learned shapes satisfy specific geometric requirements. For example, geometry-informed neural networks (GINNs) \cite{berzins2024geometry} learn shapes solely from such constraints, making them particularly powerful for learning geometries in settings where data may be sparse or unavailable. However, the inclusion of these constraints in the optimization process introduces significant challenges. Much like in physics-informed neural networks (PINNs), the optimization problems in shape learning tend to be ill-conditioned \cite{DRM}, leading to slow convergence and suboptimal accuracy when relying on first-order optimization methods like Adam. 

Second-order optimization methods, particularly the Gauss-Newton (GN) method, have recently shown great promise in accelerating convergence and achieving significantly lower errors than first-order methods in the training of the related PINNs \cite{zeinhofer2023engd, zeinhofer2024gaussnewton, zeinhofer2024position}. GN exploits the residual structure of the PDE-constrained optimization problem to mitigate the ill-conditioning that arises from high-order derivatives in the underlying differential operators \cite{de2023operator}. A key advantage of GN is its ability to resolve the mismatch between the neural network’s parameter space and the infinite-dimensional function space in which the PDE-constrained optimization problem is posed. By leveraging curvature information from the residuals, GN more effectively navigates the optimization landscape defined by the underlying functional constraints, rather than the parameter space of the neural network.
As a result, it not only converges more rapidly than first-order methods but also consistently outperforms traditional second-order optimization methods in terms of solution accuracy.

The similarities in optimization challenges between PINNs and shape learning naturally give rise to the question whether the use of the second-order methods, such as the Gauss-Newton method, which have proven effective in addressing ill-conditioning in PINNs, can similarly improve the convergence and stability in shape learning.
However, applying such methods in the context of shape learning introduces new complexities, such as formulating the geometric constraints in residual integral form and handling a moving integration domain.

\textbf{Our contributions} are:
\begin{itemize}
\item We extend the Gauss-Newton method to implicit neural surface models by reformulating key geometric losses in least-squares form.
\item We address the challenge of defining and optimizing losses over moving implicit surfaces.
\item We implement\footnote{code available at https://github.com/JamesAndrewKing/GaussNewtonShapes} a memory- and time-efficient Gauss-Newton method tailored to geometric learning tasks, enabling scalability to larger network sizes.
\item Experimentally, we demonstrate that our approach achieves significantly faster convergence and higher accuracy compared to Adam and LBFGS on benchmark shape learning tasks.
\item We provide implementation details and experimental setups to enable reproducibility and practical adoption.
\end{itemize}

\section{Background}

A powerful class of approaches in shape learning are \emph{neural fields} (NFs) \cite{Sitzmann2020ImplicitNR, essakine2025standimplicitneuralrepresentations}, also known as implicit neural representations (INRs) or coordinate-based neural networks. These models parametrize continuous functions via a multilayer perceptron
\begin{align*}
    f_{\theta}: \Omega \subseteq \mathbb{R}^d \to \mathbb{R}^b, \quad \vect{x} \mapsto \vect{y},
\end{align*}
where $\Omega$ denotes the ambient domain in which the geometry is embedded. This formulation provides smooth, memory-efficient representations with access to automatic differentiation.

\emph{Implicit neural shapes} (INSs) \cite{park2019deepsdflearningcontinuoussigned, chen2019learningimplicitfieldsgenerative} use neural fields to represent geometric structures as level sets $\Gamma(\theta) := \{f_{\theta} = 0\}$ or sublevel sets $ \Pi(\theta) := \{f_{\theta} \leq 0\}$ of a learned scalar field ($b=1$), often modeling surfaces as signed distance functions (SDFs, see \ref{sec:sdf}). This enables topological flexibility and precise surface representations. However, INSs face fundamental ill-posedness issues: 
(i) the same shape can be represented by infinitely many functions, necessitating regularization, and
(ii) multiple distinct shapes may satisfy the same design constraints, motivating generative approaches. 
To counteract this first type of ill-posedness, various strategies have been proposed, most of them requiring that the learned function behaves as an SDF.
Examples are enforcing ground-truth normals, off-surface penalties \cite{Sitzmann2020ImplicitNR}, or enforcing the eikonal equation \cite{pmlr-v119-gropp20a} \begin{align*} \norm{\nabla_{\vect{x}} f_{\theta}} = 1. \end{align*} 
This regularization stabilizes the representation by ensuring well-separated level sets and improving generalization. 
The fact that multiple shapes may satisfy given design constraints has also been approached by requiring additional constraints, like minimal surface \cite{atzmon2020saldsignagnosticlearning} or minimal volume properties.

Closely related are \emph{physics-informed neural networks} (PINNs) \cite{RAISSI2019686}, which solve PDEs by minimizing residual losses instead of ground-truth data. Given a PDE
\begin{align*}
    \mathcal{D} f = h \quad \text{in } \mathcal{X} \subseteq \Omega \text{ open}, \qquad \mathcal{B}f = g \quad \text{on } \partial \mathcal{X},
\end{align*}
with differential operator $\D$, boundary operator $\mathcal{B}$, $h \in L^2(\mathcal{X})$ and $g \in L^2(\partial\mathcal{X})$, PINNs optimize
\begin{align*}
    \frac{1}{2}\int_{\mathcal{X}} [\mathcal{D} f_{\theta} - h]^2 \, \dx + \frac{\lambda}{2}\int_{\partial \mathcal{X}} [\mathcal{B}f_{\theta} - g]^2 \, \dsigma,
\end{align*}
where $\lambda$ is a hyperparameter balancing the PDE residual and the boundary residual terms. Although this formulation typically corresponds to a well-posed problem, enforcing PDE constraints through optimization introduces significant challenges. The convergence of PINNs can be slow and unstable, especially when the underlying differential operators involve high derivatives and are thus ill-conditioned \cite{de2023operator}. Techniques such as adaptive weighting of loss terms \cite{Wang2020UnderstandingAM}, domain decomposition \cite{Moseley_2023}, or preconditioning \cite{de2023operator} have been explored to improve PINN performance, but optimization remains a key bottleneck in practical applications.

\emph{Geometry-informed neural networks} (GINNs) 
\cite{berzins2024geometry} combine ideas from INSs and PINNs to enable constraint-driven shape learning without data. Shapes are represented as level or sublevel sets and trained using a wide variety of geometric
constraints, often involving complicated differential operators. One key novelty of GINNs is the inclusion of topological losses, which allow control over topological features such as connectedness and the number of holes. 
Additionally, GINNs tackle the ill-posed nature of the problems they solve, by incorporating a generative approach, where modulation in the latent space produces a diverse solution set that satisfies the constraints, avoiding mode-collapse and enhancing design flexibility. These advancements make GINNs particularly effective for tasks requiring a range of possible design solutions while maintaining control over geometrical and topological properties.

\subsection{Losses in Shape Learning} \label{sec:losses_in_shape_learning}

In shape learning, a variety of loss functions are used to embed geometric constraints directly into the optimization process (see Table~\ref{tab:constraints}). \\
For instance, the \textit{design region loss} penalizes violations of spatial containment by discouraging the implicit shape $\Pi = \{f \leq 0\} \subseteq \Omega$ from extending outside a prescribed domain $\mathcal{E}$. \\
When part of the target interface $\I$ is known, \textit{interface losses} can be used to anchor the zero level set at the prescribed points, and \textit{normal losses} can align the implicit geometry's normals $\nabla_{\vect{x}} f / \norm{\nabla_{\vect{x}} f}$ with known target normals.\\
The \textit{eikonal loss}, enforcing $\norm{\nabla_{\vect{x}} f} = 1$, promotes $f$ to behave like a signed distance function. This regularization helps stabilize other losses and encourages better geometric fidelity, but can introduce numerical instabilities that require careful treatment \cite{Yang2023StEikST}. \\
Curvature-based losses are a natural choice for enforcing smoothness on implicit surfaces. One option, for example, is to penalize the \textit{mean curvature} $\kappa_H$ on the surface, which is computable directly from derivatives of $f$ \cite{Goldman2005CurvatureFF}. 
One may also penalize the \textit{Gaussian curvature} $\kappa_G$, which causes the curvature to vanish in at least one principal direction, favoring surfaces that bend in only one direction. A related notion is the \textit{surface strain}, which pushes both principal curvatures $\kappa_1$ and $\kappa_2$ towards zero and thus promotes overall curvature suppression. As they include second order differential operators, minimizing these curvature losses leads to especially ill-conditioned optimization problems. Another important point to consider, is that these losses are defined on the implicit surface $\Gamma = \{f = 0\}$ itself, which in turn changes during optimization. As a consequence, choosing suitable optimization schemes for such losses requires careful consideration.\\
Topological constraints can also be incorporated into the learning process, enforcing global geometric properties like connectedness or controlling the number of holes in a shape. These can be formulated as topological losses using tools like persistent homology \cite{berzins2024geometry,Clough_2022}, which dynamically identify critical data points the function must pass through to satisfy the desired topological features. Unlike conventional data terms, these losses target global shape characteristics rather than enforcing pointwise accuracy.
Although the loss function itself is formally identical to a conventional data-fitting objective (e.g. mean-squared error), the target data is recomputed at every iteration. Consequently, the optimisation effectively fits to data that moves along with the evolving, similarly to curvature-based losses.
While these topological losses don't necessarily lead to ill-conditioned optimization problems, they are expensive to evaluate and motivate optimization schemes that require few iterations.

\begin{table*}
\def\normal{ \frac{\nabla f(\vect{x})}{\norm{\nabla f(\vect{x})|}}}
    \footnotesize 
    \setlength{\tabcolsep}{6pt} 
    \renewcommand{\arraystretch}{2} 
    \centering
    \begin{tabular}{p{3.0cm}|p{4.5cm}|p{4.5cm}}
         & Function constraint & Energy functional $E(f)$ \\
        \hline
        Design region
            & $f(\vect{x}) > 0 \ \forall \vect{x} \notin \mathcal{E} $
            & $\frac{1}{2} \int_{\Omega\setminus\mathcal{E}} \left[ \operatorname{min} \{0, f(\vect{x}) \} \right]^2 \dx$ \\
        Interface
            & $f(\vect{x}) = 0 \ \forall \vect{x} \in \mathcal{I}$
            & $\frac{1}{2} \int_\mathcal{I} f^2(\vect{x}) \dsigma$ \\
        Prescribed normal
            & $\normal = \vect{n}(\vect{x}) \ \forall \vect{x} \in \mathcal{I}$
            & $\frac{1}{2} \int_{\I} \left[\scalarproduct{\normal}{\vect{n}(\vect{x})} -1 \right]^2 \dsigma$ \\
        Eikonal
            & $\norm{\nabla f(\vect{x})}=1 \ \forall \vect{x} \in \Omega$ 
            & $\frac{1}{2}\int_{\Omega} \left[\norm{\nabla f(\vect{x})}-1 \right]^2 \dx$ \\
        Mean curvature 
            & $\operatorname{div}\left(\normal\right)=0 \ \forall \vect{x} \in \Gamma$ 
            & $\frac{1}{2} \int_{\Gamma} \left[\text{div} \left( \normal \right)\right]^2 \dsigma$ \\
        Gauss curvature 
            & $\kappa_G(\vect{x})=0 \ \forall \vect{x} \in \Gamma$ 
            & $\frac{1}{2} \int_{\Gamma} \left[\kappa_G(\vect{x})\right]^2 \dsigma$ \\
        Surface strain 
            & $\kappa_1(\vect{x}) = \kappa_2(\vect{x}) = 0 \ \forall \vect{x} \in \Gamma \backslash \mathcal{I}$
            & $\frac{1}{2} \int_{\Gamma \backslash \mathcal{I}} \kappa_1^2(\vect{x}) + \kappa_2^2(\vect{x}) \dsigma$ \\
        Supervised
            & $f(\vect{x}^{(l)}) = y^{(l)}, l=1,...,N$
            & $\frac{1}{2N} \sum_{l=1}^N \left[f(\vect{x}^{(l)}) - y^{(l)} \right]^2$
    \end{tabular}
    \caption{
    Geometric constraints and their corresponding energy functionals.}
    \label{tab:constraints}
\end{table*}

\subsection{Mathematical Framework}
\label{sec:mathematical_setting}

We consider a general geometric optimization problem where the objective is to minimize a functional \( E_1: \mathcal{H} \to \mathbb{R} \) over a Hilbert space \( \mathcal{H} \), subject to geometric constraints \( E_i(f) = 0 \) for \( i = 2, \dots, n \). These constraints are termed \textit{energy functionals}, and we convert them into soft constraints by solving the unconstrained optimization problem
\begin{align*}
    \min_{f \in \mathcal{H}} E(f),
\end{align*}
where the total energy functional is given by
\begin{align*}
    E(f) := \sum_{i=1}^n \lambda_i E_i(f),
\end{align*}
with weight parameters \( \lambda_i \).

To solve this problem, we use a family of parametric functions, typically neural networks, \( \mathcal{M} := \{f_{\theta}: \theta \in \Theta\} \subseteq \mathcal{H} \), and optimize over the parameter space \( \Theta \subseteq \mathbb{R}^p \) with the optimization problem
\begin{align*}
    \min_{\theta \in \Theta} \mathcal{L}(\theta),
\end{align*}
where the loss function \( \mathcal{L}(\theta) := E(f_{\theta}) \) corresponds to the energy functional for the parameterized function \( f_{\theta} \).

We assume the differentiability of the map \( P: \Theta \to \mathcal{M} \), where \( \theta \mapsto f_{\theta} \), leading to the notion of the generalized tangent space \( T_{\theta} \mathcal{M} = \text{range}(DP(\theta)) = \text{span} \{ \partial_{\theta_1} f_{\theta}, \dots, \partial_{\theta_p} f_{\theta} \} \).

Focusing on energies of the form
\begin{align*}
    E_i(f) = \frac{1}{2} \int_{\Omega_i} [R_i(f)]^2 \, \dx = \frac{1}{2} \| R_i(f) \|_{L^2(\Omega_i)}^2,
\end{align*}
where \( R_i: \mathcal{H} \to L^2(\Omega_i) \) is a twice Fréchet differentiable residual functional and \( \Omega_i \subseteq \Omega \), we express these constraints in \( L^2 \)-energy form. The corresponding losses are given by
\begin{align}
    L_i(\theta) = \frac{1}{2} \int_{\Omega_i} [r_i(\theta)]^2 \, \dx = \frac{1}{2} \| r_i(\theta) \|_{L^2(\Omega_i)}^2,
    \label{eq:loss_residual_form}
\end{align}
where \( r_i = R_i \circ P \), and the gradient of the loss with respect to \( \theta \) is related to the residual by
\begin{align}
    \partial_{\theta_j} r_i(\theta) = DR_i(f_{\theta}) \partial_{\theta_j} f_{\theta}.
    \label{eq:relation_gradient_residual_penalty}
\end{align}

\subsection{Optimization Algorithms}
\label{sec:optimization_algorithms}
Given the challenges in shape learning, first order optimization methods may be insufficient, particularly due to the ill-conditioned nature of embedding differential constraints into the learning process. 
While network sizes in deep learning frequently lie in the millions, making second order optimization methods difficult to use efficiently, networks in shape learning are much smaller and models with only a few thousand parameters can often suffice in capturing intricate surface details.

A widely used gradient-based method is \textit{gradient descent}, defined as
\begin{align}
    \theta_{k+1} = \theta_k - \eta \nabla_{\theta} \L(\theta_k),
\end{align}
where $\theta_k$ is the current parameter, $\eta$ is the learning rate, and $\nabla_{\theta} \L(\theta_k)$ is the gradient of the loss function. While simple, gradient descent can struggle in non-convex settings, particularly if the loss landscape is poorly conditioned.

\textit{Newton's method} enhances gradient descent by incorporating curvature information from the Hessian matrix
\begin{align}
    \theta_{k+1} = \theta_k - \eta [H_{\theta}^{\L}]^{-1} \nabla_{\theta} \L(\theta_k),
\end{align}
resulting in locally quadratic convergence requiring fewer iterations \cite{Bjrck1996NumericalMF}. However, computing the Hessian and solving a linear system at each step is computationally expensive. A more efficient approach is the \textit{Gauss-Newton method}, 
\begin{align}
    \theta_{k+1} = \theta_k - \eta \A_{\text{gn}}^{-1}(\theta_k) \nabla_{\theta} \L(\theta_k),
\end{align}
where $\A_{\text{gn}}$ captures the positive semi-definite (PSD) part of the full Hessian \( H_{\theta}^{\L} \) neglecting the second-order derivatives of the residuals, and is given by
\begin{align}
\A_{\text{gn}}(\theta) := \sum_{i=1}^n \lambda_i \int_{\Omega_i} \nabla_{\theta} r_i(\theta) 
    \nabla_{\theta} r_i(\theta)^T \, \dx \approx H_{\theta}^{\L}.
\end{align}
This method is particularly useful when the residuals are approximately linear or small \cite{Bjrck1996NumericalMF}. For numerical stability, the \textit{Levenberg-Marquardt algorithm} \cite{Bjrck1996NumericalMF} combines Gauss-Newton with a regularization term, allowing for a more robust approach that adapts to the problem's complexity.

\textit{LBFGS}, a quasi-Newton method, reduces the computational cost by approximating the inverse Hessian using a limited history of gradients and parameter updates. This memory-efficient approach is particularly advantageous for large-scale problems where computing or storing the full Hessian is prohibitive.
Finally, \textit{Adam} is an adaptive first-order method that balances momentum and per-parameter learning rates, making it suitable for noisy or sparse gradient problems.

\subsection{Natural Gradient Descent}

Optimization is a primary challenge in shape learning, much like it for PINNs \cite{de2023operator}. As a result, methods such as natural gradient descent, which are derived from optimizing in function space and have proven effective for PINNs \cite{zeinhofer2023engd, zeinhofer2024gaussnewton, zeinhofer2024position}, can also offer advantages in this context.
To keep the exposition clear, we focus our analysis on a single energy functional of the form
\begin{align} 
E(f) = \frac{1}{2} \norm{R(f)}^2_{L^2(\Omega)}. \label{eq:functional_loss} 
\end{align} 
Combining multiple constraints can be handled analogously by treating the total loss as a linear combination of individual terms.


The idea of natural gradient descent \cite{Amari1998NaturalGW} is to design an optimization algorithm in $\mathcal{H}$ and from it derive a corresponding algorithm in $\Theta$ which retains desirable properties and ensures that updates move optimally in parameter space with respect to the function space geometry.
For this, consider the scheme
\begin{align}
    f_{k+1} = f_k + \eta d_k, 
    \label{eq:function_space_scheme}
\end{align}
with learning rate $\eta$ and update direction $d_k$. We choose 
\begin{align}
    d_k := -T_k^{-1}(DE(f_k)),
    \label{eq:function_space_update_direction}
\end{align}
where $DE(f_k)\in \mathcal{H}^*$ is the Fréchet derivative at $f_k$ and $T_k: \mathcal{H} \rightarrow \mathcal{H}^*$ is a positive definite linear operator describing how the gradient step is transformed. \\
It can be shown \cite{zeinhofer2024position} that the parameter space algorithm that mimics the evolution of \eqref{eq:function_space_scheme} up to first order in the Riemannian metric induced by $T_k$, is given by 
\begin{align}
    \theta_{k+1} = \theta_{k} - \eta G(\theta_k)^\dagger \nabla_{\theta} L(\theta_k),
    \label{eq:parameter_space_scheme}
\end{align}
where $G(\theta_k)^\dagger$ denotes the pseudoinverse of the Gramian matrix
\begin{align}
    G(\theta_k)_{ij} = \scalarproduct{\I^{-1}T_k \partial_{\theta_i} f_{\theta_k}}{\partial_{\theta_j} f_{\theta_k}}_\H,
    \label{eq:gramian_matrix}
\end{align}
and $\I$ is the Riesz isomorphism. Conversely, updating the parameters according to \eqref{eq:parameter_space_scheme} amounts to moving according to \eqref{eq:function_space_scheme} in function space and then orthogonally projecting onto $T_{\theta_k}\mathcal{M}$.

\subsection{Gauss-Newton Natural Gradient Descent}
\label{sec:gauss_newton_natural_gradient_descent}
Having discussed the connection between function space and parameter space optimization algorithms in the previous section, we can now aim to develop sensible gradient-based methods in function space.\\
We do this by choosing a certain $d_k$ in \eqref{eq:function_space_scheme}, ideally
\begin{align}
    d_k &= \arg \min_{d \in \H} E(f_k + d) \nonumber \\
    &= \arg \min_{d \in \H} \frac{1}{2} \norm{R(f_{k} + d)}^2_{L^2(\Omega)}.
    \label{eq:function_space_ideal_direction}
\end{align}
This nonlinear least squares problem can be approached in multiple ways using the form \eqref{eq:function_space_update_direction}, but given the nature of our energy functional as a squared norm over a residual, it makes sense to develop methods based on this form. \\
Importantly, also in function space, this problem is generally ill-conditioned due to the properties of the differential operators in $R$, especially when they are of high order or nearly singular. This motivates the use of second-order methods, which can better handle such ill-conditioning by incorporating curvature information.

Gauss-Newton methods aim to solve \eqref{eq:function_space_ideal_direction}, by solving a locally linearized version leading to the linear least squares problem 
\begin{align}
    d_k = \arg \min_{d \in \H} \frac{1}{2} \norm{R(f_{k}) + DR(f_k)d}^2_{L^2(\Omega)},
    \label{eq:function_space_linear_least_squares}
\end{align}
at every iteration. The solution to \eqref{eq:function_space_linear_least_squares} is given by the normal equations
\begin{align*}
    d_k = -[DR(f_k)^* DR(f_k)]^{-1}(DR(f_k)^*R(f_k))
\end{align*}
which we can write in the form of \eqref{eq:function_space_update_direction} with 
\begin{align*}
    T_k = \mathcal{I} \circ [DR(f_k)^* DR(f_k)].
\end{align*}

Inserting this into \eqref{eq:gramian_matrix} and recalling the chain rule \eqref{eq:relation_gradient_residual_penalty} gives us the Gramian matrix

\begin{align*}
    G(\theta_k)_{ij} &= \scalarproduct{\partial_{\thetathin_i} r(\theta_k)}{\partial_{\thetathin_j} r(\theta_k)}_{L^2(\Omega)} \nonumber\\
    &= \A_{\text{gn}}(\theta_k)_{ij},
\end{align*}
which is just the Gauss-Newton preconditioner in parameter space. Therefore $L^2$-Gauss-Newton Natural Gradient Descent in function space corresponds to the Gauss-Newton method in parameter space modulo the projection onto $T_{\theta}\M$. 

\section{Methods}

Most constraints listed in Table \ref{tab:constraints} can be translated to loss functions in residual form \eqref{eq:loss_residual_form} very easily, but strictly speaking, this form assumes constant integration domains. In the following section, we show how for losses over implicit surfaces, one can approximate the induced gradient flow by resampling on the surface at every step. The subsequent sections provide details for implementing the Gauss-Newton method in the context of shape learning. 

\subsection{Losses over Moving Surfaces}
\label{sec:losses_over_moving_surfaces}
When approximating losses involving integrals over an implicit surface $\Gamma(\theta) := \{q_{\theta} = 0\}$ where $q_{\theta}$ is a smooth function parametrized by $\theta$ (for example the model $f_{\theta}$ itself), we must account for the fact that both the integrand and the integration domain depend on the parameters $\theta$. For a generic surface loss of the form
\begin{align*}
    L_{\text{surf}}(\theta) = \frac{1}{2}\int_{\Gamma(\theta)} [r(\theta)]^2 \, \mathrm{d}\sigma,
\end{align*} 
its gradient can be computed using the Reynolds transport theorem for surfaces \cite{Delhaye1974JumpCA}
\begin{align}
    \partial_{\theta_i} L_{\text{surf}}(\theta) = &\int_{\Gamma(\theta)} r(\theta) \, \partial_{\theta_i} r(\theta) \nonumber\\
    &+r(\theta)\scalarproduct{\nabla_{\Gamma}r(\theta)}{\vect{v}}\nonumber\\ 
    &+\frac{1}{2} [r(\theta)]^2 \, \mathrm{div}_{\Gamma}(\vect{v}) \, \mathrm{d}\sigma,
    \label{eq:grad_moving_surface_loss}
\end{align}
where $\nabla_\Gamma$ denotes the surface gradient and $\vect{v}$ is the velocity field describing how the level set $\Gamma(\theta)$ deforms as $\thetathin_i$ varies. The evolution of the implicit function $q_{\theta}$ under this velocity field is governed by the \textit{level set advection equation} \cite{osher2004level},
\begin{align*}
    \partial_{\thetathin_i} q_{\theta} + \scalarproduct{\vect{v}}{\nabla_\vect{x} q_{\theta}} = 0.
\end{align*}
As the evolution of $q_{\theta}$ only depends on the normal component of $\vect{v}$, we are free to set the tangential component zero. With this choice, the second term in \eqref{eq:grad_moving_surface_loss} vanishes.\\
In practice, it is common to ignore the dependency of the surface on $\theta$ and instead resample points on $\Gamma(\theta)$ at each iteration, effectively treating the surface as fixed. This approximation neglects the third term in \eqref{eq:grad_moving_surface_loss} and can be justified when $[r(\theta)]^2 \ll 1$, which is a valid assumption in the vicinity of the minimizer. However, common practice is to make this assumption along the entire optimization path, thus assuming that the term is small enough from initialization. 
Surface resampling can impact optimizers like LBFGS, which rely on past gradients and assume a consistent loss landscape over time. When sampling points on the surface change at each iteration, the gradients no longer correspond to the same objective, potentially leading to instabilities. Gauss-Newton, on the other hand, operates solely on the current residual and its local linearization, making it more robust in the presence of such shifting domains. As a result, for any such loss that involves changing sampling points, such as smoothness or topological losses, the Gauss-Newton method is a better choice than LBFGS.

\subsection{Implementation}
\label{sec_implementation}
When implementing any of the optimization algorithms in Section \ref{sec:optimization_algorithms}, the appearing integrals in the loss function have to be approximated numerically. This also extends to the Gauss-Newton Gramian matrix $\A_{\text{gn}}(\theta_k)$. As the integration domains in shape learning can become very complicated, we resort to Monte Carlo quadrature, but the following approach holds for any kind of numerical quadrature. \\
For this, Let $\{\vect{x}_j^i, \ j=1,...,N_i\}$ be uniform sampling points in $\Omega_i$. With this we can approximate
\begin{align}
    \L(\theta_k) &= \sum_{i=1}^n \frac{\lambda_i}{2} \int_{\Omega_i} [r_i(\theta_k)]^2 \dx \nonumber\\
    &\approx \sum_{i=1}^n\frac{\Tilde{\lambda}_i}{2N_i} \sum_{j=1}^{N_i} [r_i(\theta_k, \vect{x}_j^i)]^2
    \label{eq:monte_carlo_quadrature_loss}\\
    \A_{\text{gn}}(\theta_k) &= \sum_{i=1}^n \lambda_i \int_{\Omega_i} \nabla_{\theta} r_i(\theta_k) \nabla_{\theta} r_i(\theta_k)^T \, \dx \nonumber\\
    &\approx \sum_{i=1}^n \frac{\Tilde{\lambda}_i}{N_i} J_i^T J_i,
    \label{eq:monte_carlo_quadrature_gn}
\end{align}
where $(J_i)_{ml} = \partial_{\thetathin_l} r_i(\theta_k, \vect{x}_m^i)$ is the Jacobian in $\R^{N_i\times p}$ and $\Tilde{\lambda}_i := \lambda_i \abs{\Omega_i}$ can be seen as loss reweighting.
We compute the loss and Gramian terms efficiently using \texttt{torch.func.vmap()} and \texttt{torch.func.jacrev()} for each residual individually. Once assembled, the Gauss-Newton update becomes
\begin{align*}
    \delta_k &= \arg \min_{\mathbf{v} \in \mathbb{R}^p} \|(\A_{\text{gn}}(\theta_k) + \epsilon I) \mathbf{v} - \nabla_{\theta} \mathcal{L}(\theta_k)\|^2 \\
    \theta_{k+1} &= \theta_k - \eta \delta_k,
\end{align*}
where $\epsilon$ is a regularization parameter. This least-squares problem can be solved efficiently on a GPU using \texttt{torch.linalg.lstsq()} with QR decomposition, preferably in double precision for better accuracy. The learning rate $\eta$ can be adapted at each step using a line-search. Similarly to \cite{zeinhofer2024gaussnewton}, we found that sampling $\eta$ equidistantly in log-space worked well enough while retaining efficiency. 

\subsection{Scaling}
The approach presented in Section \ref{sec_implementation} works very well for small parameter sizes $p$. However, for larger networks this implementation would become inefficient as the dense matrix solve scales with $\mathcal{O}(p^3)$. The following sections address measures that can be taken to scale up the Gauss-Newton method to larger networks.

\subsubsection{Matrix-Free Solvers}
\label{sec:matrix_free_solvers}
To improve on the cubic scaling of the matrix solve, we can exploit the structure of the Gauss-Newton Gram matrix for a more efficient approach using Krylov methods \cite{zeinhofer2024gaussnewton}.
Equation \eqref{eq:monte_carlo_quadrature_gn} gives rise to the decomposition
\begin{align}
    \A_{\text{gn}}(\theta_k) = J^T J,
    \label{eq:JJ_T}
\end{align}
where $J\in \mathbb{R}^{N \times p}$ is the Jacobian of the full residual
\begin{align*}
    \mathbf{r} = \left(\sqrt{\frac{\Tilde{\lambda}_1}{N_1}} \mathbf{r}_1^T \mid \cdots \mid \sqrt{\frac{\Tilde{\lambda}_n}{N_n}} \mathbf{r}_n^T \right)^T \in \mathbb{R}^N, (\mathbf{r}_i)_j = r_i(\theta_k, \vect{x}_j^i)
\end{align*}
and $N = \sum_{i=1}^n N_i$. Using the decomposition \eqref{eq:JJ_T}, we can compute matrix-vector products $\A_{\text{gn}}(\theta_k) \mathbf{v} = J^T(J \mathbf{v})$ in $\mathcal{O}(pN)$ relying on the rectangular matrix $J$ which is much more tractable than $\A_{\text{gn}}(\theta_k)$. Note that computing $J$ is an intermediate step of the computation of the loss gradient, making it readily available. The regularization can then be applied to the matrix-vector product. This allows the use of iterative solvers to solve the linear system $\A_{\text{gn}}(\theta_k)\vect{v} = \nabla_{\theta} L(\theta_k)$, like Krylov subspace methods including GMRES or the CG-Method, which rely solely on matrix-vector products. The convergence speed of such iterative methods depends on the condition number of $\A_{\text{gn}}(\theta_k)$ which motivates preconditioning the iterative solver using for example (Block-)Jacobi or randomized Nyström preconditioning \cite{frangella2021randomizednystrompreconditioning}.

\subsubsection{Scaling via the Woodbury Identity}
An alternative scaling strategy \cite{guzmáncordero2025improvingenergynaturalgradient} for deep neural networks can be achieved using the Woodbury \textit{push-through} identity. Taking the gradient of equation \eqref{eq:monte_carlo_quadrature_loss} gives us
\begin{align*}
    \nabla_{\theta} \mathcal{L}(\theta_k) &= \sum_{i=1}^n \frac{\Tilde{\lambda}_i}{N_i} \sum_{j=1}^{N_i} r_i(\theta_k, \vect{x}_j^i) \nabla_{\theta} r_i(\theta_k, \vect{x}_j^i) = J^T \mathbf{r}.
\end{align*}
The regularized Gauss-Newton update becomes
\begin{align*}
    \theta_{k+1} = \theta_k + (\epsilon I + J^TJ)^{-1} J^T \mathbf{r}.
\end{align*}
Using the push-through identity
\begin{align*}
    (\epsilon I + UV)^{-1} U = U (\epsilon I + VU)^{-1},
\end{align*}
we obtain
\begin{align*}
    \theta_{k+1} = \theta_k + J (\epsilon I + J J^T)^{-1} \mathbf{r}.
\end{align*}
This allows the computational complexity to scale with $N$ instead of $p$, making it feasible for deep networks when $N$ is sufficiently small. For summing over parameters when calculating $J J^T$, we recommend batching per layer for memory efficiency. 

Apart from the ideas presented here, it is possible to efficiently scale the Gauss-Newton method using Kronecker-Factorized Approximate Curvature \cite{dangel2024kroneckerfactoredapproximatecurvaturephysicsinformed} or techniques from randomized linear algebra  \cite{guzmáncordero2025improvingenergynaturalgradient, hellmuth2024preserving}.

\section{Results}

We evaluate the performance of our method across four distinct shape learning tasks: minimal surfaces, developable surfaces, general implicit shape reconstruction from data, and a simple engineering design problem. Each setting imposes different primary geometric constraints: including several types of curvatures, topology, and fidelity to surface point clouds with ground truth normals. Across all experiments, we compare Gauss-Newton against Adam and LBFGS, measuring quantitative losses and qualitative geometric accuracy. The results consistently show that Gauss-Newton yields faster convergence, lower errors, and superior surface quality.

\subsection{Plateau's Problem}
\label{sec:minimal_surfaces}

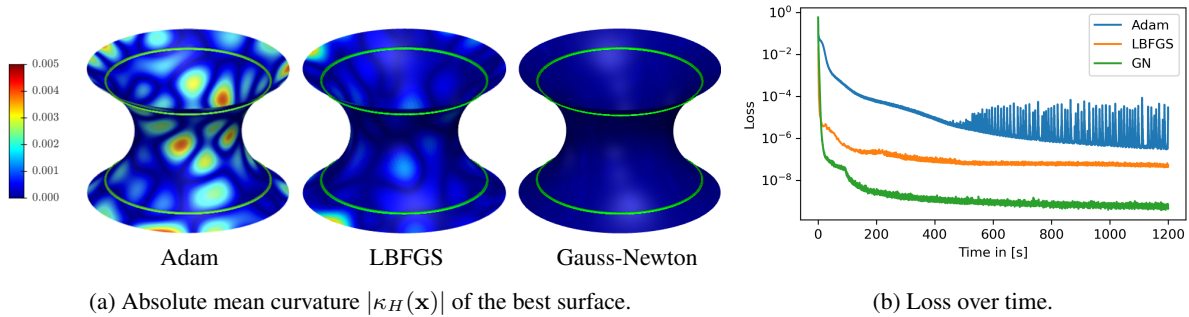
\begin{figure}[H]
    \centering
    \begin{subfigure}[t]{0.58\textwidth}
        \flushleft
        \begin{tikzpicture}
            \node[inner sep=0] (img) {\includegraphics[width=\linewidth]{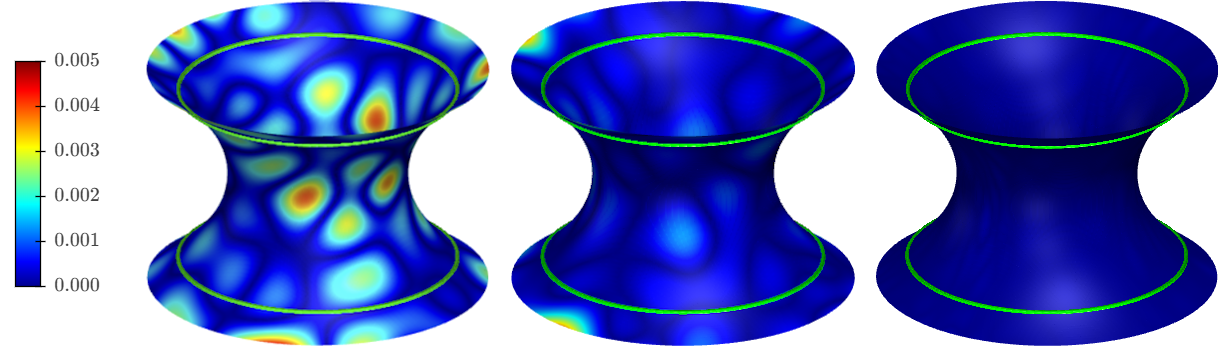}};
            \node[below=2pt of img.south west, anchor=north west, xshift=0.21\linewidth] {\footnotesize Adam};
            \node[below=2pt of img.south west, anchor=north west, xshift=0.50\linewidth] {\footnotesize LBFGS};
            \node[below=2pt of img.south west, anchor=north west, xshift=0.76\linewidth] {\footnotesize Gauss-Newton};
        \end{tikzpicture}
        \caption{Absolute mean curvature $\abs{\kappa_H(\vect{x})}$ of the best surface.}
        \label{fig:heatmaps_catenoid}
    \end{subfigure}
    \begin{subfigure}[t]{0.38\textwidth}
        \flushright
        \includegraphics[width=\linewidth]{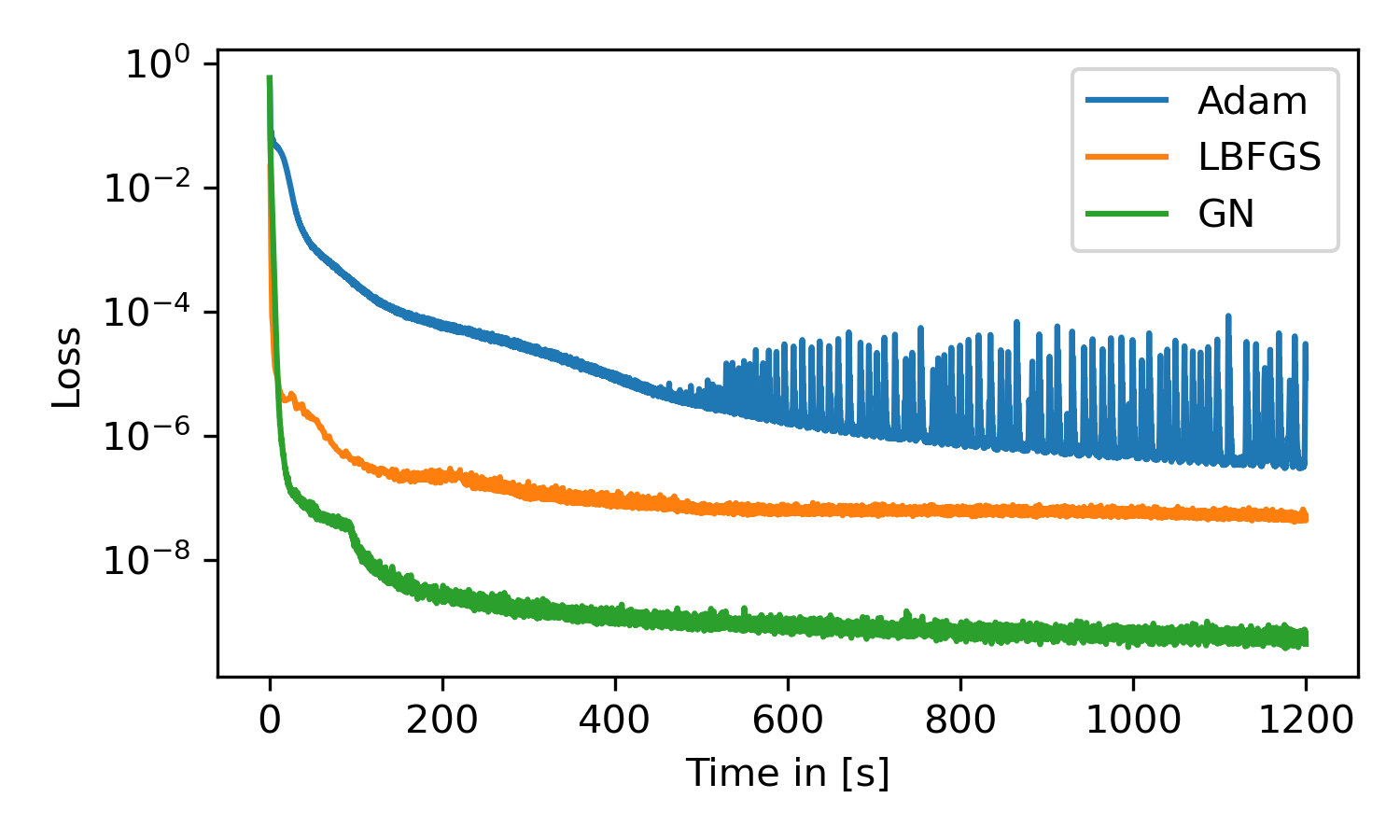}
        \caption{Loss over time.}
        \label{fig:errors_catenoid}
    \end{subfigure}
    \caption{
    Plateau's problem on the catenoid geometry with mean curvature and interface loss. The prescribed interface is highlighted in green in (a). The chamfer divergence to the ground truth is additionally shown in Figure \ref{fig:chamfer_catenoid}.
    }
    \label{fig:catenoid_full}
\end{figure}

\begin{figure}[H]
    \centering
    \begin{subfigure}[t]{0.58\textwidth}
        \flushleft
        \begin{tikzpicture}
            \node[inner sep=0] (img) {\includegraphics[width=\linewidth]{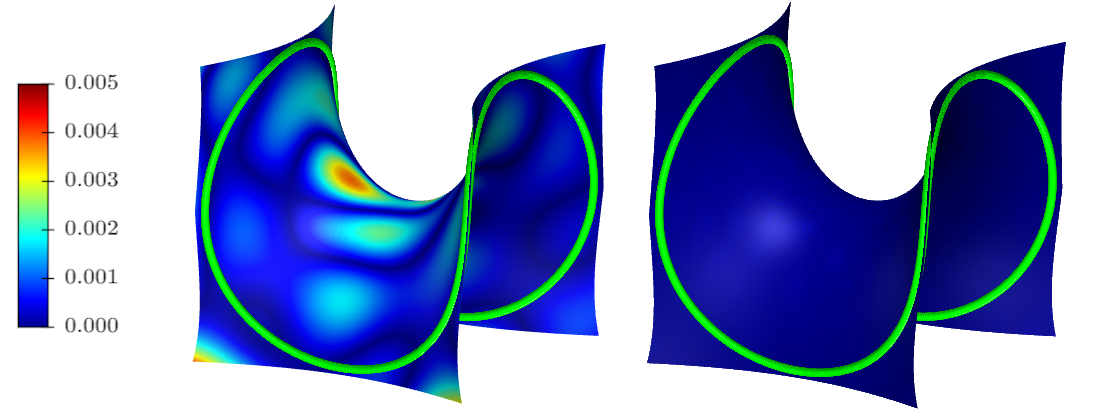}};
            \node[below=2pt of img.south west, anchor=north west, xshift=0.30\linewidth] {\footnotesize Adam};
            \node[below=2pt of img.south west, anchor=north west, xshift=0.66\linewidth] {\footnotesize Gauss-Newton};
        \end{tikzpicture}
        \caption{Absolute mean curvature $\abs{\kappa_H(\vect{x})}$ of the best surface.}
        \label{fig:heatmaps_enneper}
    \end{subfigure}
    \begin{subfigure}[t]{0.38\textwidth}
        \flushright
        \includegraphics[width=\linewidth]{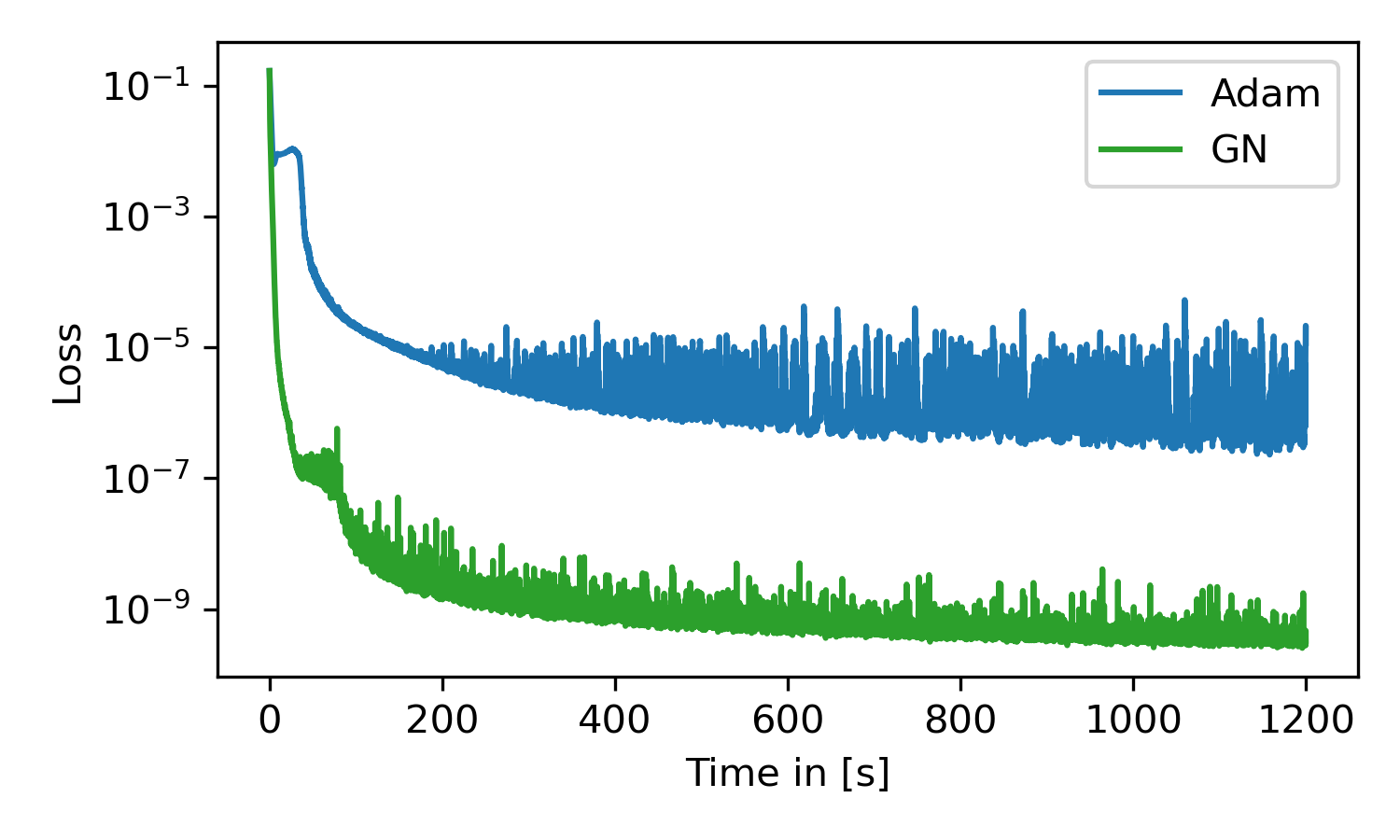}
        \caption{Loss over time.}
        \label{fig:errors_enneper}
    \end{subfigure}
    \caption{
    Plateau's problem on the Enneper minimal surface geometry with mean curvature and interface loss. The prescribed interface is highlighted in green in (a). LBFGS becomes unstable and is omitted. The chamfer divergence to the ground truth is additionally shown in Figure \ref{fig:chamfer_enneper}.
    }
    \label{fig:enneper_full}
\end{figure}

In shape optimization, a key problem is learning a smooth surface that connects to a given interface. We focus on learning minimal surfaces as a special case. The problem of finding a minimal surface enclosed by a given Jordan curve is called \textit{Plateau's Problem}. Minimal surfaces locally minimize surface area \cite{Nitsche1975VorlesungenM}, which is equivalent to having zero mean curvature.
We consider the catenoid and Enneper surface—two classical solutions with known analytic forms—as test cases. These examples allow us to assess how well different optimization methods minimize mean curvature while fitting to the prescribed interface. For error measurement, we use the Chamfer Divergence $CD_1(P, Q)$ between the model surface $P$ and the true surface $Q$ (see Section \ref{sec:chamfer_divergence}). Figures \ref{fig:heatmaps_catenoid} to \ref{fig:errors_enneper} show the results of training for 20 minutes, in which Gauss-Newton completed $\sim 8000$ iterations, LBFGS $\sim 9000$ and Adam $\sim 17000$.
Our results show that Gauss-Newton achieves lower losses, errors, and faster convergence than Adam and LBFGS.

\subsection{Developable Surfaces}
\label{sec:developable_surfaces}
Another class of geometrically constrained surfaces is \emph{developable surfaces}, characterised by having zero Gaussian curvature $\kappa_G$ at every point. Developable surfaces can be flattened onto a plane without distortion. This constraint appears naturally in applications such as sheet metal design and architectural geometry.
To assess performance, we train our model to represent the mantle of a truncated cone, a canonical example of a smooth developable surface. Similar to the minimal surface experiments, we enforce an interface constraint and minimize a geometric objective - here the Gauss curvature. \\
The qualitative and quantitative results mirror those observed for minimal surfaces: Gauss-Newton consistently achieves lower errors and faster convergence than the first- and quasi-second-order baselines. This highlights the effectiveness of Gauss-Newton in learning globally consistent surfaces under curvature-based constraints.

\begin{figure}[H]
    \centering
    \begin{subfigure}[t]{0.58\textwidth}
        \flushleft
        \begin{tikzpicture}
            \node[inner sep=0] (img) {\includegraphics[width=\linewidth]{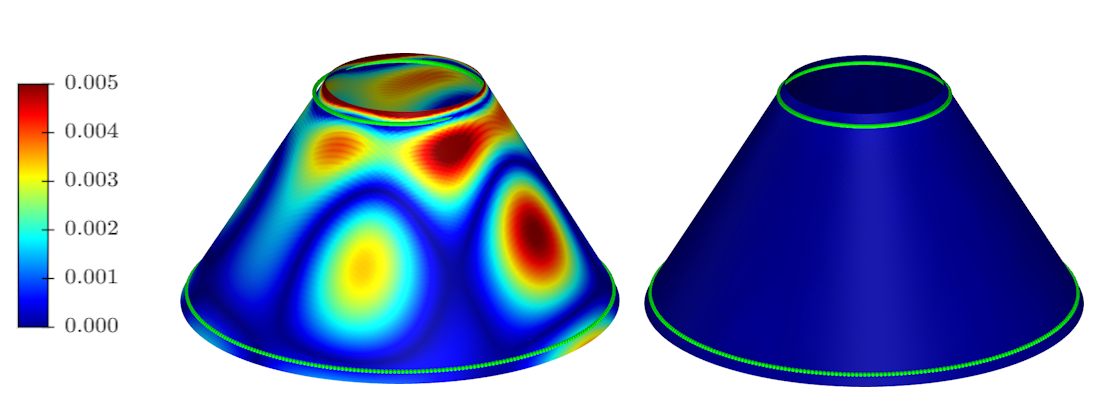}};
            \node[below=2pt of img.south west, anchor=north west, xshift=0.32\linewidth] {\footnotesize Adam};
            \node[below=2pt of img.south west, anchor=north west, xshift=0.67\linewidth] {\footnotesize Gauss-Newton};
        \end{tikzpicture}
        \caption{Absolute Gauss curvature $\abs{\kappa_G(\vect{x})}$ of the best surface.}
        \label{fig:heatmaps_cone}
    \end{subfigure}
    \begin{subfigure}[t]{0.38\textwidth}
        \centering
        \includegraphics[width=\linewidth]{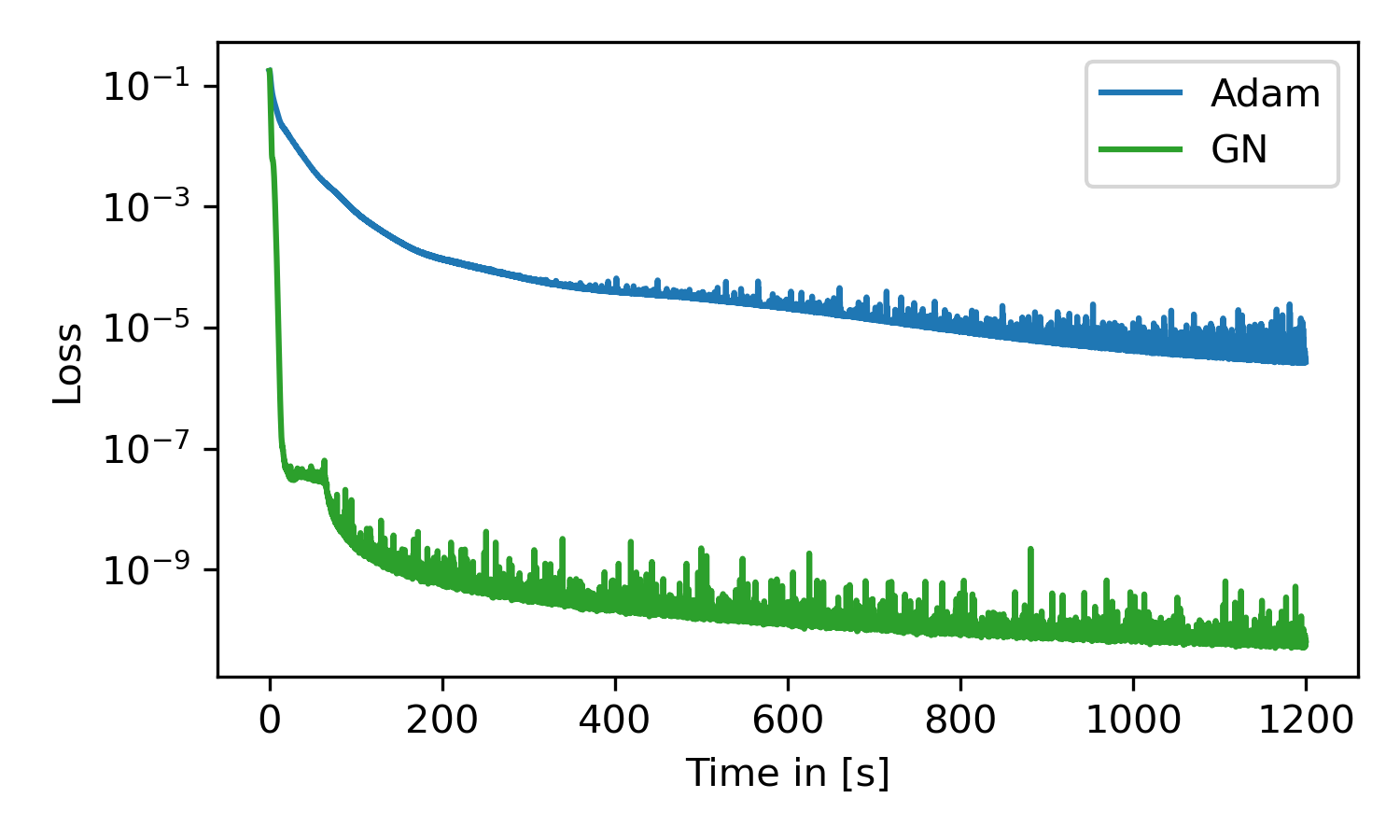}
        \caption{Loss over time.}
        \label{fig:errors_cone}
    \end{subfigure}
    \caption{Developable surface problem on the truncated cone geometry with Gauss curvature and interface loss. The prescribed interface is highlighted in green in (a). LBFGS becomes unstable and is omitted.}
    \label{fig:cone_full}
\end{figure}

\subsection{Implicit Neural Shapes}
\label{sec:implicit_neural_shapes}
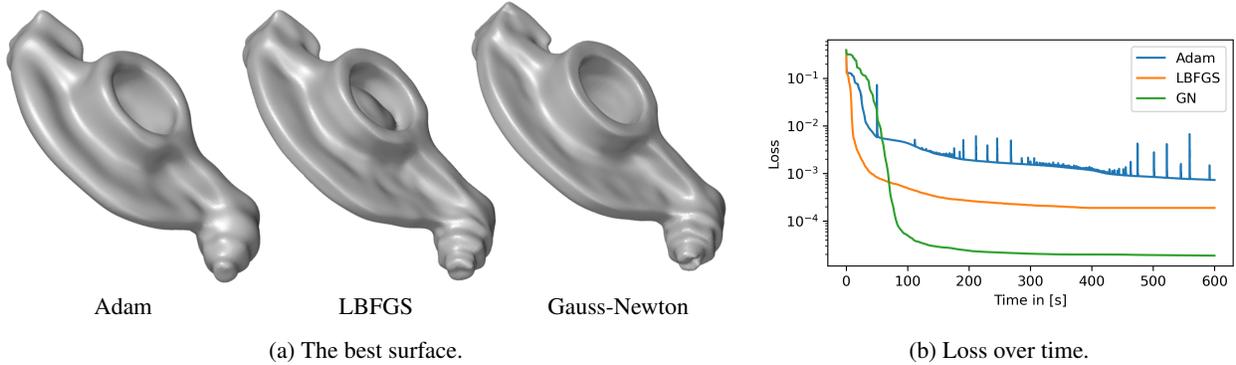
\begin{figure}[H]
    \begin{subfigure}[t]{0.58\textwidth}
        \flushleft
        \begin{tikzpicture}
            \node[inner sep=0] (img) {\includegraphics[width=\linewidth]{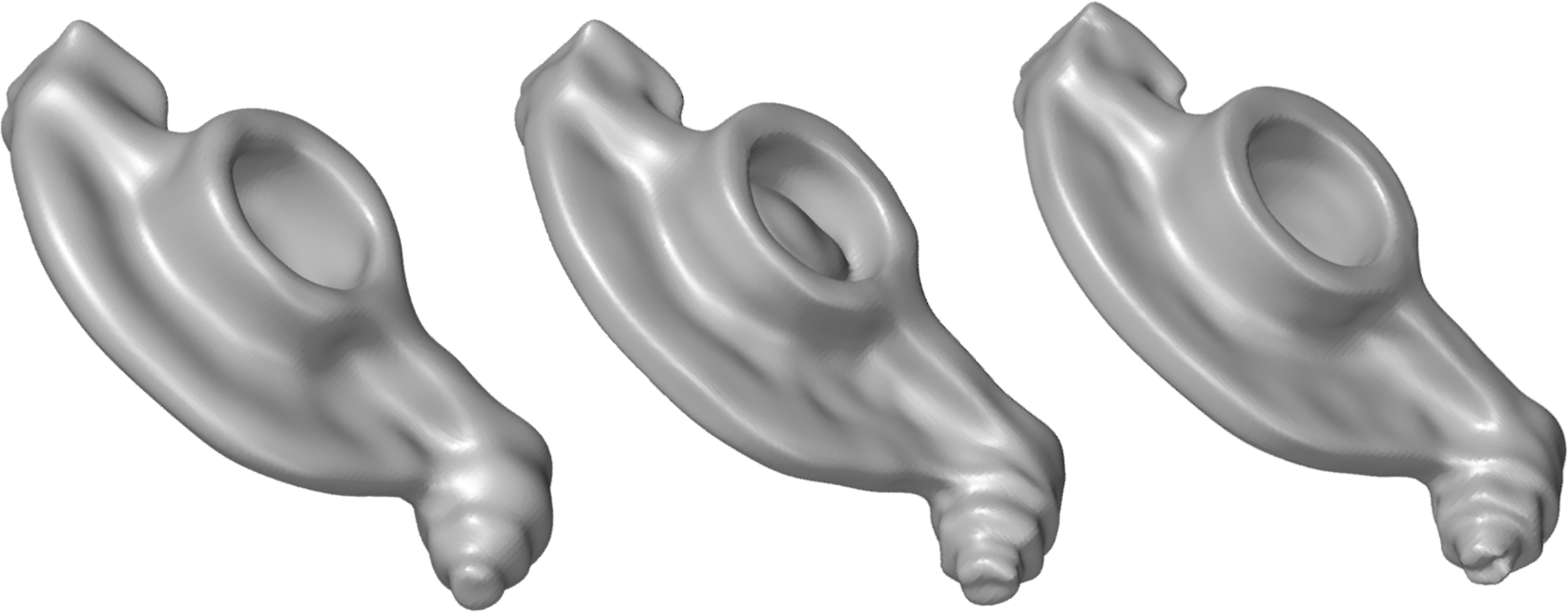}};
            \node[below=2pt of img.south west, anchor=north west, xshift=0.11\linewidth] {\footnotesize Adam};
            \node[below=2pt of img.south west, anchor=north west, xshift=0.45\linewidth] {\footnotesize LBFGS};
            \node[below=2pt of img.south west, anchor=north west, xshift=0.74\linewidth] {\footnotesize Gauss-Newton};
        \end{tikzpicture}
        \caption{The best surface.}
        \label{fig:rockerarm_comparison}
    \end{subfigure}
    \hfill
    \begin{subfigure}[t]{0.40\textwidth}
        \flushright
        \includegraphics[width=\linewidth]{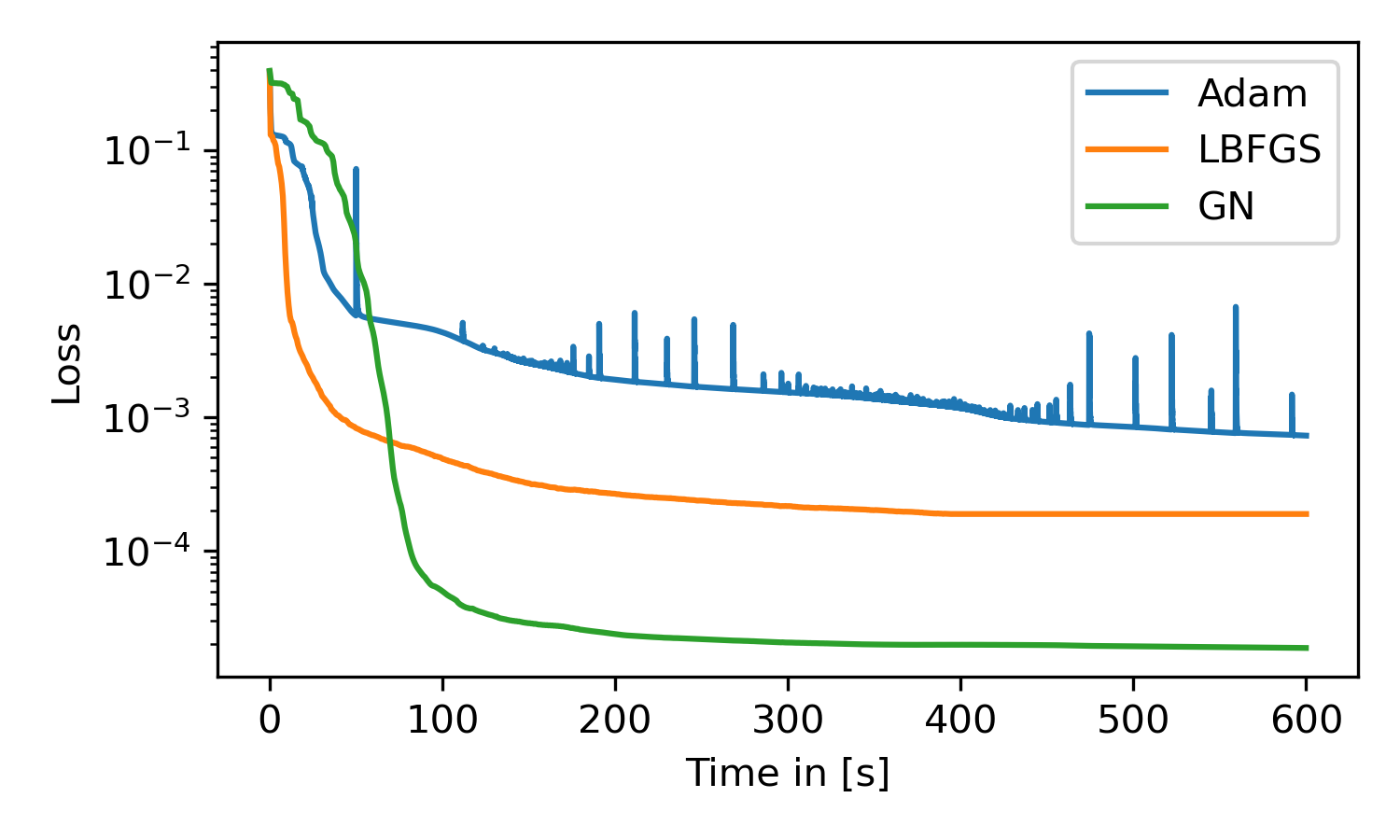}
        \caption{Loss over time.}
        \label{fig:errors_rockerarm}
    \end{subfigure}

    \caption{Implicit neural shape problem on the rocker arm geometry with interface and normal loss.
    }
    \label{fig:rockerarm_full}
\end{figure}

\begin{figure}[H]
    \begin{subfigure}[t]{0.58\textwidth}
        \flushleft
        \begin{tikzpicture}
            \node[inner sep=0] (img) {\includegraphics[width=\linewidth]{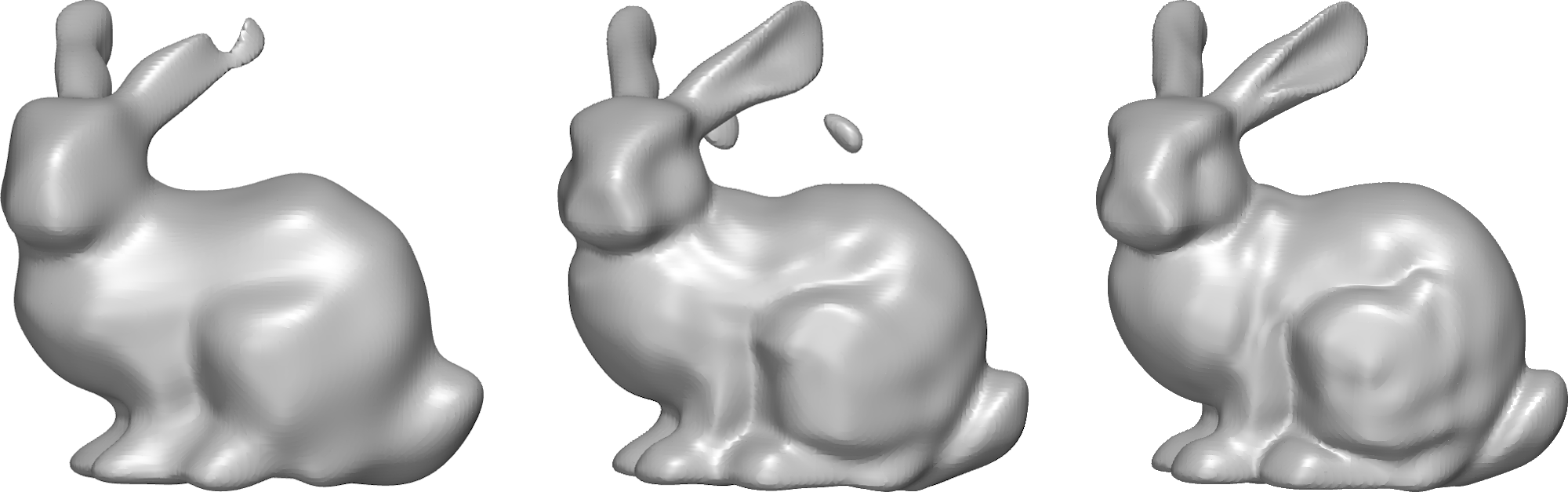}};
            \node[below=2pt of img.south west, anchor=north west, xshift=0.11\linewidth] {\footnotesize Adam};
            \node[below=2pt of img.south west, anchor=north west, xshift=0.45\linewidth] {\footnotesize LBFGS};
            \node[below=2pt of img.south west, anchor=north west, xshift=0.74\linewidth] {\footnotesize Gauss-Newton};
        \end{tikzpicture}
        \caption{The best surface.}
        \label{fig:bunnies_comparison}
    \end{subfigure}
    \hfill
    \begin{subfigure}[t]{0.40\textwidth}
        \flushright
        \includegraphics[width=\linewidth]{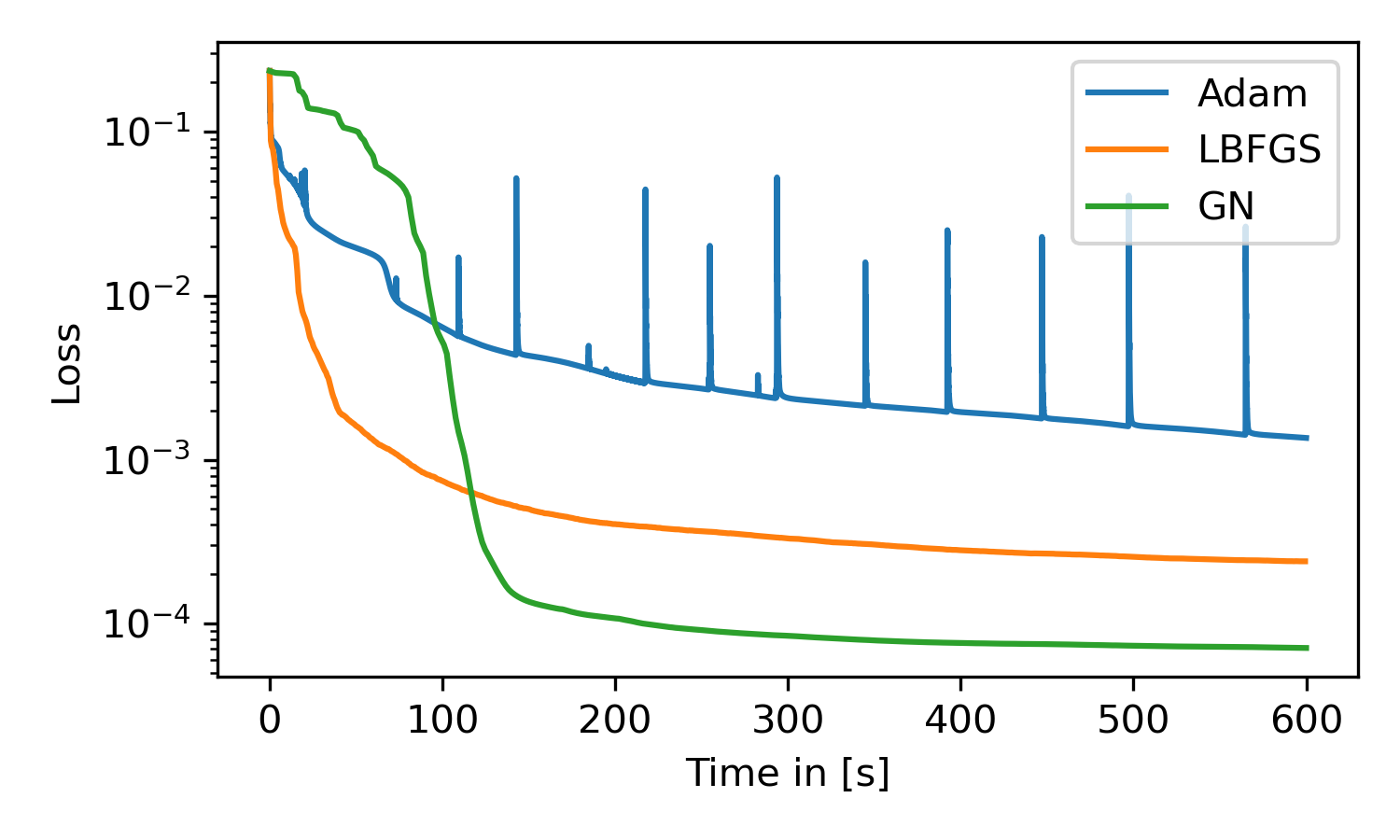}
        \caption{Loss over time.}
        \label{fig:errors_bunny}
    \end{subfigure}

    \caption{Implicit neural shape problem on the Stanford bunny geometry with interface and normal loss.
    }
    \label{fig:bunny_full}
\end{figure}
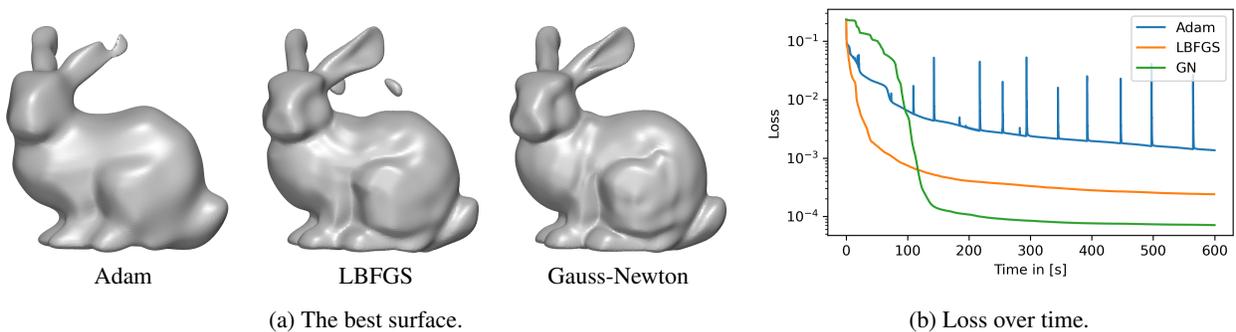

We evaluate our method on two standard benchmark shapes—the Rocker Arm and the Stanford Bunny—using a loss that enforces closeness to the ground truth surface points, encourages alignment with the prescribed normals, and is regularized by an eikonal penalty. As shown in Figures \ref{fig:errors_rockerarm} and \ref{fig:errors_bunny}, Gauss-Newton consistently achieves the lowest loss across both examples, outperforming Adam and LBFGS. The qualitative results in Figures \ref{fig:rockerarm_comparison} and \ref{fig:bunnies_comparison} further demonstrate that Gauss-Newton produces reconstructions with finer geometric detail and better visual agreement with the target meshes. 
We also found Gauss-Newton to be particularly effective in data-sparse settings, achieving similar mesh quality and loss reduction when trained on only 20-30\% of the available ground truth data. In these experiments, we randomly discarded a large portion of the surface data, using only the remaining points for training. Despite this significant reduction in data, the resulting meshes and loss curves closely matched those from the full data experiments, with the added benefit of faster Gauss-Newton steps. This highlights the usefulness of Gauss-Newton for applications where only limited data is available for reconstruction.

\subsection{Jet Engine Bracket}
\label{sec:jet_engine_bracket}

The last use-case considers the design of a jet engine bracket \cite{GE, berzins2024geometry}. Conceptually, this geometry-informed neural network task is similar to the first two experiments, but with more numerous and computationally costly constraints. These include an eikonal regularization, a non-trivial design region enclosed by a watertight mesh, and six cylindrical interfaces with inward-facing normals illustrated in Figure \ref{fig:simjeb_def}. Additionally, we minimize \emph{surface strain}, which is another curvature-based quantity, but different to the previous use-cases, it is not zero everywhere for the (unknown and possibly non-unique) optimal shape. We expect this loss contribution to be severely ill-conditioned.
\\
Lastly, \emph{connectedness} enforces that the shape does not have disconnected components -- a prerequisite for the structural integrity of a part. Despite being a discrete property, \emph{persistent homology} (PH) can relax this (and other topological properties) into differentiable quantities that can be optimized continuously.
In essence, Morse theory describes that, as we sweep the level $t\in\mathbb{R}$, the connected components of the sublevel-set $\{f_{\theta}<t\}$ are born, merged, split or destroyed at precisely the critical points of $f_{\theta}$. PH identifies these critical points $\{\vect{x}_i\}$, and the desired topological changes can be achieved by optimizing the values $f_{\theta}(\vect{x}_i)$ accordingly.
Once the points and target values are determined, the loss itself can be implemented as a simple data-loss in our framework.
Notably, the PH computation itself is rather expensive (around a second), while differentiating through $f_{\theta}(\vect{x}_i)$ is trivial. This motivates using as few iterations as possible.

Similar to the first two problems, we initialize with (i.e., pretrain on) an approximate SDF of the design region to stabilize training. This is known to help shape learning \cite{Atzmon2020sal, atzmon2020saldsignagnosticlearning}.
Furthermore, due to the complexity of the problem, the evaluation points change in each iteration, both due to sampling a fixed domain (e.g., interface, design region) or due to the moving domain itself (e.g., the iso-surface and critical points in PH).

While fixed sampling points are generally assumed for the convergence proofs of nonlinear least squares algorithms, resampling does work in practice if the Hessian approximation is adapted to this resampling. As we point out in Section~\ref{sec:losses_over_moving_surfaces}, this is the case for Gauss-Newton. 

\begin{figure}
    \centering
    \begin{subfigure}[t]{0.58\textwidth}
        \flushleft
        \begin{tikzpicture}
            \node[inner sep=0] (img) {\includegraphics[width=\linewidth]{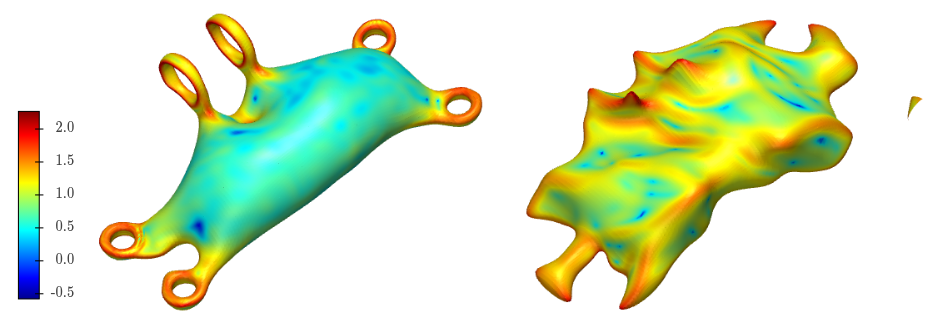}};
            \node[below=2pt of img.south west, anchor=north west, xshift=0.18\linewidth] {\footnotesize Gauss-Newton};
            \node[below=2pt of img.south west, anchor=north west, xshift=0.67\linewidth] {\footnotesize Adam};
        \end{tikzpicture}
        \caption{Surface strain of the best surface.}
        \label{fig:ginn_noclip_comparison}
    \end{subfigure}
    \begin{subfigure}[t]{0.40\textwidth}
        \flushright
        \includegraphics[width=\linewidth]{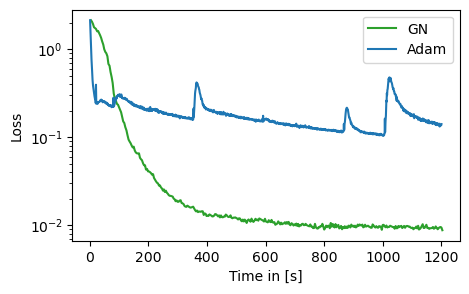}
        \caption{Loss over time.}
        \label{fig:errors_ginn_noclip}
    \end{subfigure}
    \caption{Jet engine bracket problem \emph{without clipping} the ill-conditioned strain. The additional constraints include interface, interface normal, design region, connectedness, and eikonal regularization,  illustrated in Figure \ref{fig:simjeb_def}. 
    (a) The surface strain $\kappa_1^2(\vect{x}) + \kappa_2^2(\vect{x})$ in $\log_{10}$-scale for the best shapes identified by each optimizer. 
    Adam fails to converge to a feasible shape as also quantified by the total loss over time in (b).
    }
    \label{fig:ginn_noclip_full}
\end{figure}

\begin{figure}
    \centering
    \begin{subfigure}[t]{0.58\textwidth}
        \flushleft
        \begin{tikzpicture}
            \node[inner sep=0] (img) {\includegraphics[width=\linewidth]{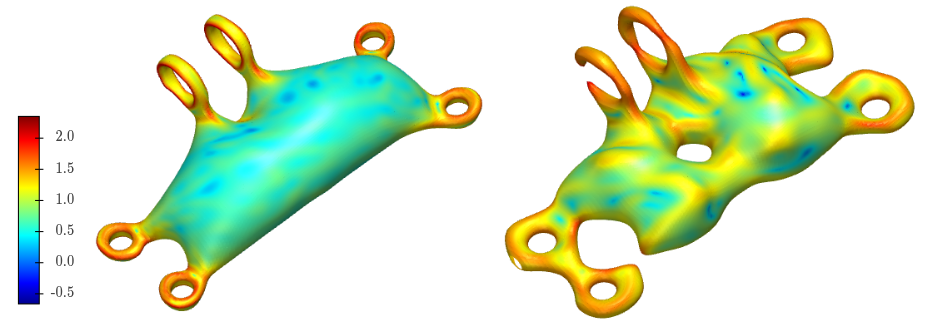}};
            \node[below=2pt of img.south west, anchor=north west, xshift=0.18\linewidth] {\footnotesize Gauss-Newton};
            \node[below=2pt of img.south west, anchor=north west, xshift=0.67\linewidth] {\footnotesize Adam};
        \end{tikzpicture}
        \caption{Surface strain of the best surface.}
        \label{fig:ginn_clip_comparison}
    \end{subfigure}
    \begin{subfigure}[t]{0.40\textwidth}
        \flushright
        \includegraphics[width=\linewidth]{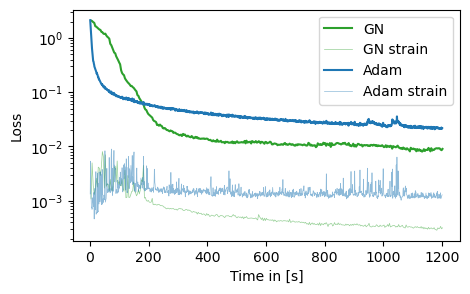}
        \caption{Loss over time.}
        \label{fig:errors_ginn_clip}
    \end{subfigure}
    \caption{Jet engine bracket problem \emph{with clipping} the ill-conditioned strain.  The additional constraints include interface, interface normal, design region, connectedness, and eikonal regularization,  illustrated in Figure \ref{fig:simjeb_def}. 
    (a) The surface strain $\kappa_1^2(\vect{x}) + \kappa_2^2(\vect{x})$ in $\log_{10}$-scale for the best shapes identified by each optimizer. 
    (b) Total loss over time. Additionally plotting the ill-conditioned strain contribution reveals that Adam fails to optimize the strain at all. While clipping stabilizes and helps Adam significantly, it is still outperformed by Gauss-Newton.
    }
    \label{fig:ginn_clip_full}
\end{figure}

The results in Figure \ref{fig:ginn_noclip_full} show
good convergence for Gauss-Newton, while Adam plateaus early and does not produce a reasonable shape.
One of the reasons for this is that curvature and strain vary over several orders of magnitude. As a remedy, we clip the maximum value of the strain to 1000, which stabilizes both optimizers, as shown in Figure \ref{fig:ginn_clip_full}. While clipping helps Adam significantly and it produces an approximately feasible shape, it is still outperformed by GN. Additionally plotting the strain loss term over the optimization time reveals that Adam entirely fails to optimize this ill-conditioned loss.



\section{Conclusion, Limitations and Future Work}

In this paper, we discussed how incorporating geometric constraints into the learning process of implicit neural shapes leads to the notion of shape learning \cite{berzins2024geometry}. As in the case of PINNs, many of the associated optimization challenges stem from the ill-conditioning of differential operators, motivating the use of preconditioned methods \cite{de2023operator}. Designing such methods benefits from reasoning in function space first, where the problem is often better behaved, and then translating to parameter space, where the nonlinear network parametrization induces complex loss landscapes. This perspective motivates the use of the Gauss-Newton method for shape learning \cite{zeinhofer2024gaussnewton}. We validated this through experiments on minimal surfaces, developable surfaces, and implicit neural shape fitting, where Gauss-Newton consistently outperformed Adam and LBFGS. Moreover, we showed how to scale Gauss-Newton to large networks, broadening its usability.

While promising, our approach has limitations.
Gauss-Newton remains expensive for very large models despite our improvements, and further work on randomized and low-rank variants is well justified. Balancing many competing constraints also remains a challenge, which could be addressed by combining Gauss-Newton with augmented Lagrangian techniques used in GINNs \cite{berzins2024geometry}. Finally, the ability to handle losses over moving domains opens up promising directions in neural topology optimization, where optimization challenges in parameter space are a known bottleneck \cite{sanu2024neuraltopologyoptimizationgood}.
Overall, this work contributes toward better optimization strategies for shape learning, with potential impact across physics-based modelling, surrogate learning, shape optimization, and computational design.

\section{Funding declaration}

This research did not receive any specific grant from funding agencies in the public, commercial, or not-for-profit sectors.

\bibliographystyle{elsarticle-num} 
\bibliography{refs}

\newpage
\appendix
\onecolumn

\section{Signed Distance Function}
\label{sec:sdf}

Let $\Omega \subseteq \R^d$ be a domain. A signed distance function (SDF) $f: \Omega \to \mathbb{R}$ of a shape $\Pi \subseteq \Omega$ provides the (signed) distance from a query point $\vect{x} \in \Omega$ to the closest point on the boundary $\Gamma := \partial \Pi$
\begin{align*}
    f(\vect{x}) =
    \begin{cases}
        \text{d}(\vect{x}, \Gamma) & \text{if } \vect{x} \notin \Pi \text{ (outside the shape)}, \\
        -\text{d}(\vect{x}, \Gamma) & \text{if } \vect{x} \in \Pi \text{ (inside the shape)}.
    \end{cases}
\end{align*}

A point $\vect{x}\in \Omega$ belongs to the \textit{medial axis} of $\Pi$ if its closest boundary point is not unique. The gradient of an SDF satisfies the eikonal equation $\|\nabla_{\vect{x}} f\| = 1$ everywhere except on the medial axis, where the gradient is undefined. In shape learning, we often aim to learn a shape by approximating its SDF $f$ by a neural network $f_{\theta}$.

\section{Surface Sampling}
\label{sec:surface_sampling}
This section describes the surface sampling techniques used in the minimal and developable surface experiments (Sections~\ref{sec:minimal_surfaces} and~\ref{sec:developable_surfaces}). Efficient sampling on implicit surfaces is crucial both for evaluating losses and computing convergence errors. While fixed domains or sublevel sets ${f_{\theta} \leq 0}$ can often be sampled via rejection methods, sampling \emph{on} an implicitly defined surface without a parametrization requires root-finding techniques. Although the methods we employ do not guarantee uniformity, a sufficiently large and uniform distribution of initial points in space typically yields an approximately uniform surface distribution in practice. Post-processing methods can further improve uniformity if needed \cite{Chopp2001SomeIO}.

\emph{Binary search sampling} offers an efficient way to find points on an implicit surface. Starting from two points $\vect{x}_{\text{low}}$ and $\vect{x}_{\text{high}}$ such that $f_{\theta}(\vect{x}_{\text{low}}) < 0 < f_{\theta}(\vect{x}_{\text{high}})$, it iteratively bisects the interval and updates the bounds based on the sign of $f_{\theta}$ at the midpoint. This method avoids gradient computations and converges logarithmically, making it well-suited for generating rough surface samples quickly, though it becomes slower when high accuracy is required.

\emph{Orthogonal Newton sampling} \cite{Chopp2001SomeIO} refines points onto the surface using gradients of the implicit function. Starting from an initial point $\vect{x}_0$, it iteratively updates via
\begin{equation*}
\vect{x}_{n+1} = \vect{x}_{n} - \frac{f_{\theta}(\vect{x}_{n})} {\norm{\nabla_\vect{x} f_{\theta}(\vect{x}_{n})}}\vect{n},
\end{equation*}
where $\vect{n}$ is the unit normal. This update, which orthogonally projects points onto the surface, can be derived from a first-order expansion of $f_{\theta}$ and equivalent to applying Newton’s method along the gradient direction. With locally quadratic convergence and gradient information computed via autodifferentiation, it enables highly accurate and efficient surface sampling. A step-size cap improves stability.

\section{Chamfer Divergence}
\label{sec:chamfer_divergence}

This metric is used to evaluate convergence in the minimal surface experiments (see Section~\ref{sec:minimal_surfaces}). Let $P,Q \subseteq \R^d$ be two shapes. We define the one-sided chamfer divergence\footnote{The common definition is without the square root, but we include it for interpretability.}
\begin{align}
    CD_1(P,Q) := \sqrt{\frac{1}{\abs{Q}}\sum_{\vect{x}\in Q}\min_{\vect{y}\in P}\norm{\vect{x}- \vect{y}}^2}.
    \label{eq:chamfer_divergence}
\end{align}
While commonly used, this metric is sensitive to sampling density: unequal point counts can distort results, even for nearly identical shapes. Since our experiments involve very small divergences, we mitigate this by projecting points from one surface onto the other using Orthogonal Newton Sampling (see Section \ref{sec:surface_sampling}) to establish accurate correspondences.

\begin{figure}[H]
    \begin{subfigure}[t]{0.48\textwidth}
        \flushleft
        \includegraphics[width=\linewidth]{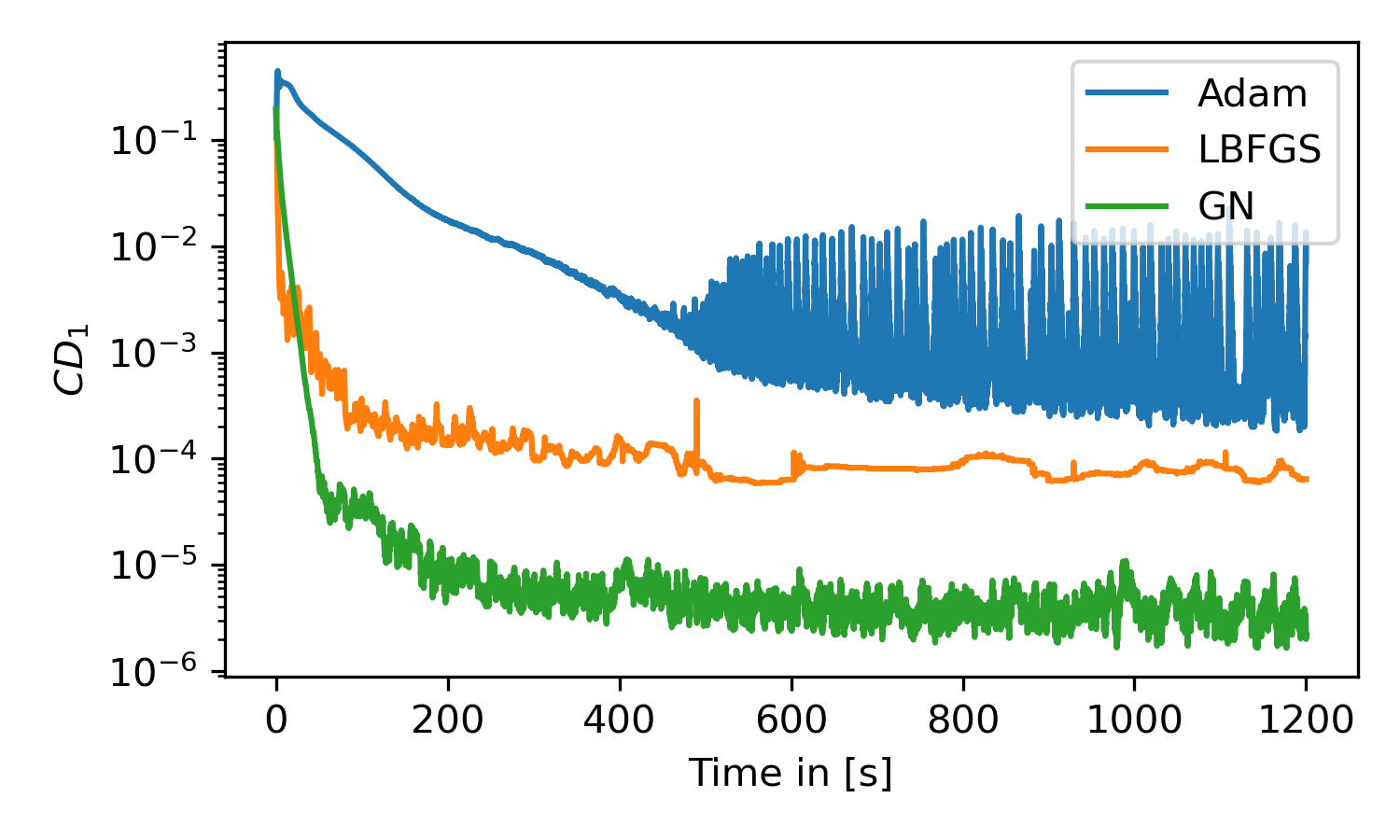}
        \caption{Catenoid (cf. Figure \ref{fig:catenoid_full}).}
        \label{fig:chamfer_catenoid}
    \end{subfigure}
    \hfill
    \begin{subfigure}[t]{0.48\textwidth}
        \flushright
        \includegraphics[width=\linewidth]{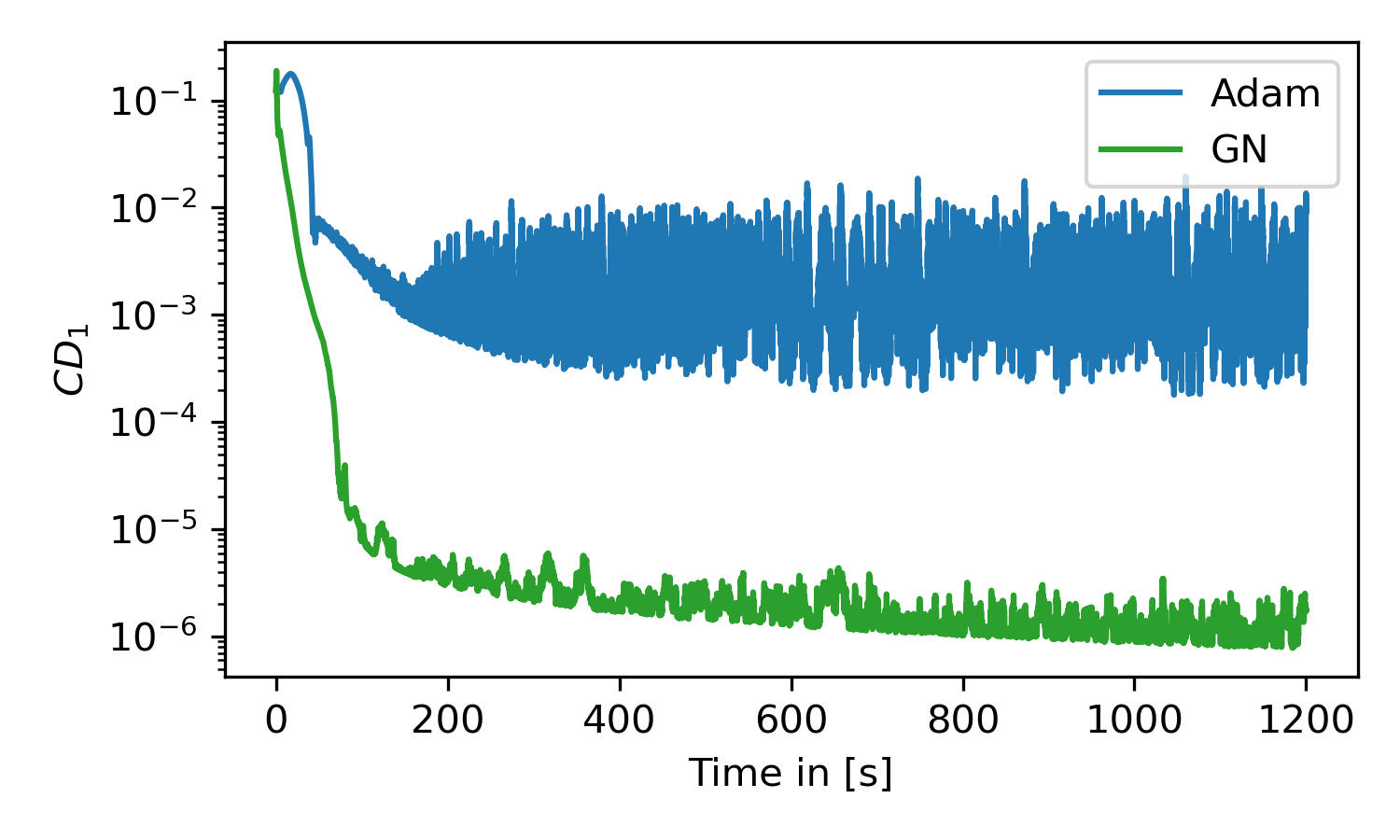}
        \caption{Enneper surface (cf. Figure \ref{fig:enneper_full}).}
        \label{fig:chamfer_enneper}
    \end{subfigure}
    \caption{Chamfer divergences over time quantifying the deviation from the known ground truth. These relate similarly to the training loss curves with Gauss-Newton converging both faster and to lower values.}
    \label{fig:chamfer_full}
\end{figure}

\section{Jet Engine Bracket Constraints Visualized}

\begin{figure}[H]
    \centering
    \includegraphics[width=0.4\linewidth]{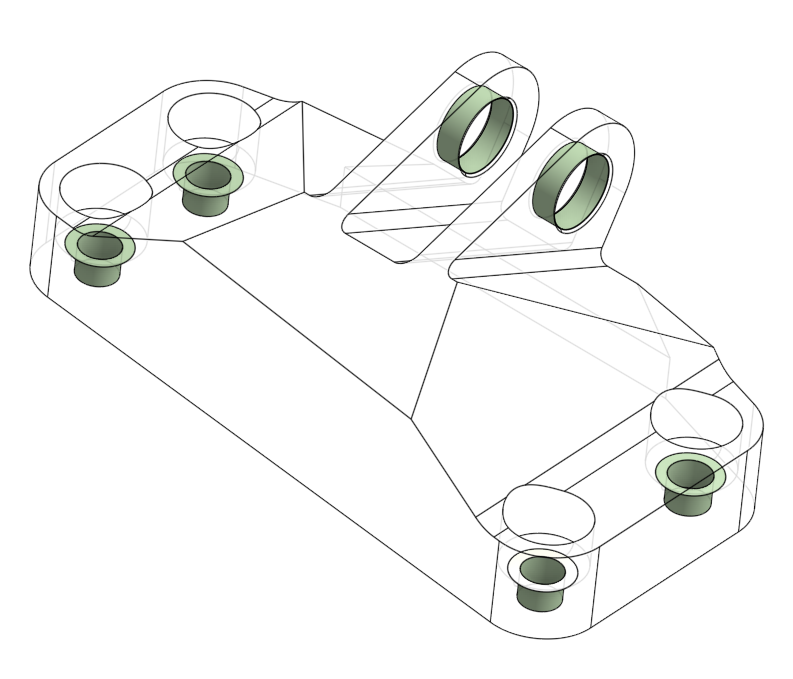}
    \caption{The constraints of the jet engine bracket problem illustrated. Any generated shape must be contained in the sketched design region and attach to the six cylindrical interfaces in green.}
    \label{fig:simjeb_def}
\end{figure}

\section{Hyperparameters and Model Architectures}
Across experiments \ref{sec:minimal_surfaces}, \ref{sec:developable_surfaces} and \ref{sec:implicit_neural_shapes}, all loss terms were assigned a weight of $1$ to ensure fair comparisons and reflect realistic out-of-the-box performance. The only exception was the eikonal regularization term, for which we performed a small hyperparameter sweep over values $\{10^{-i}, i=0 \dots 4\}$ separately for each optimizer, because Adam profits from a slightly larger regularization.
\\
For the minimal surface (\ref{sec:minimal_surfaces}) and developable surface (\ref{sec:developable_surfaces}) experiments, we used a fully-connected network architecture of size $[3, 32, 32, 1]$ with tanh activation, while the implicit neural shapes (\ref{sec:implicit_neural_shapes}) experiments used an architecture of size $[3, 32, 32, 32, 1]$ with sine activation. Learning rates were slightly tuned per method and task: for Gauss-Newton we used a fixed learning rate of $10^{-1}$ for the minimal and developable surface experiments, and a line search scheme for INSs. In both experiments we use a regularization of $10^{-6}$. Adam used learning rates of $10^{-3}$ for minimal surfaces, $10^{-4}$ for developable surfaces, and $10^{-3}$ with cosine annealing for INSs. For consistency, we pretrain each model on the approximate SDF of a shape that is topologically similar to the desired shape, i.e. the $xy$-plane (Enneper Surface), a cylinder (Catenoid and Cone) or a sphere (Rockerarm and Bunny).
The instability of LBFGS on the Enneper Surface and Cone examples can be explained with the surface samples moving along the optimization. This instability is not apparent in the Catenoid experiment, presumably because the initialization as a cylinder is significantly closer to the true shape as for the Cone and Enneper Surface.
For the jet engine bracket experiment (\ref{sec:jet_engine_bracket}), we use a NN again sized $[3, 32, 32, 1]$, but with the wavelet implicit neural network (WIRE) \cite{saragadam2023wire} architecture following the original GINN implementation \cite{berzins2024geometry}. 
We first pre-train for 3000 iterations on batches of 32768 samples of the approximate SDF of the design region. This is known to stabilize shape learning \cite{Atzmon2020sal, atzmon2020saldsignagnosticlearning}.
For the training, we weight the residuals as follows: strain $10^{-4}$, eikonal $1$, interface $10$, interface normals $10$, design region $10^3$, and connectedness $10^3$. The Gauss-Newton optimizer uses line search and a small $10^{-7}$ regularization for stability. 

\end{document}

%% file: macrosetup.tex
\newcommand{\A}{\mathbb{A}}

\newcommand{\D}{\mathcal{D}}

\newcommand{\R}{\mathbb{R}}

\newcommand{\M}{\mathcal{M}}
\renewcommand{\L}{\mathcal{L}}
\newcommand{\I}{\mathcal{I}}
\renewcommand{\H}{\mathcal{H}}
\newcommand{\dx}{\text{d}\textbf{x}}
\newcommand{\dsigma}{\text{d}\sigma}

\newcommand{\abs}[1]{\left\lvert#1\right\rvert}
\newcommand{\scalarproduct}[2]{\left\langle #1, #2 \right\rangle}
\newcommand{\vect}[1]{\mathbf{#1}}
\let\thetathin\theta
\renewcommand{\theta}{\boldsymbol{\thetathin}}
\newcommand\restr[2]{{
  \left.\kern-\nulldelimiterspace
  #1
  \vphantom{\big|}
  \right|_{#2}
  }}

\DeclarePairedDelimiter\norm{\lVert}{\rVert}

\renewcommand{\epsilon}{\ensuremath\varepsilon}
